\documentclass[final]{cvpr}
\usepackage{times}
\usepackage{epsfig}
\usepackage{graphicx}
\usepackage{amsmath}
\usepackage{amssymb}

\usepackage[pagebackref=false,breaklinks=true,colorlinks,bookmarks=false]{hyperref}

\usepackage{subfigure}
\usepackage{float}
\usepackage{bbm}
\usepackage{booktabs}
\usepackage{pifont}
\newcommand{\cmark}{\ding{51}}%
\newcommand{\xmark}{\ding{55}}%
\usepackage{multirow}
\usepackage{flushend}
\usepackage{dblfloatfix}
\usepackage{threeparttable}

\def\cvprPaperID{0000} 
\def\confYear{arxiv 2021}
\begin{document}

\title{Fully Convolutional Networks for Panoptic Segmentation}

\author{Yanwei Li$^{1}$\quad Hengshuang Zhao$^{2}$\quad Xiaojuan Qi$^{3}$\quad Liwei Wang$^{1}$\quad \\
Zeming Li$^{4}$\quad Jian Sun$^{4}$\quad Jiaya Jia$^{1,5}$  \\[0.2cm]
The Chinese University of Hong Kong$^{1}$\quad University of Oxford$^{2}$\\
The University of Hong Kong$^{3}$\quad MEGVII Technology$^{4}$ \quad SmartMore$^{5}$
}

\maketitle

\begin{abstract}
   In this paper, we present a conceptually simple, strong, and efficient framework for panoptic segmentation, called Panoptic FCN. Our approach aims to represent and predict foreground things and background stuff in a unified fully convolutional pipeline. In particular, Panoptic FCN encodes each object instance or stuff category into a specific kernel weight with the proposed kernel generator and produces the prediction by convolving the high-resolution feature directly. With this approach, instance-aware and semantically consistent properties for things and stuff can be respectively satisfied in a simple generate-kernel-then-segment workflow. Without extra boxes for localization or instance separation, the proposed approach outperforms previous box-based and -free models with high efficiency on COCO, Cityscapes, and Mapillary Vistas datasets with single scale input. Our code is made publicly available at \href{https://github.com/Jia-Research-Lab/PanopticFCN}{https://github.com/Jia-Research-Lab/PanopticFCN}.\footnote{Part of the work was done in MEGVII Research.}
\end{abstract}

\section{Introduction}
\begin{figure}[th!] \label{fig:intro}
\centering
\subfigure[Separate representation]{
\includegraphics[width=0.45\linewidth]{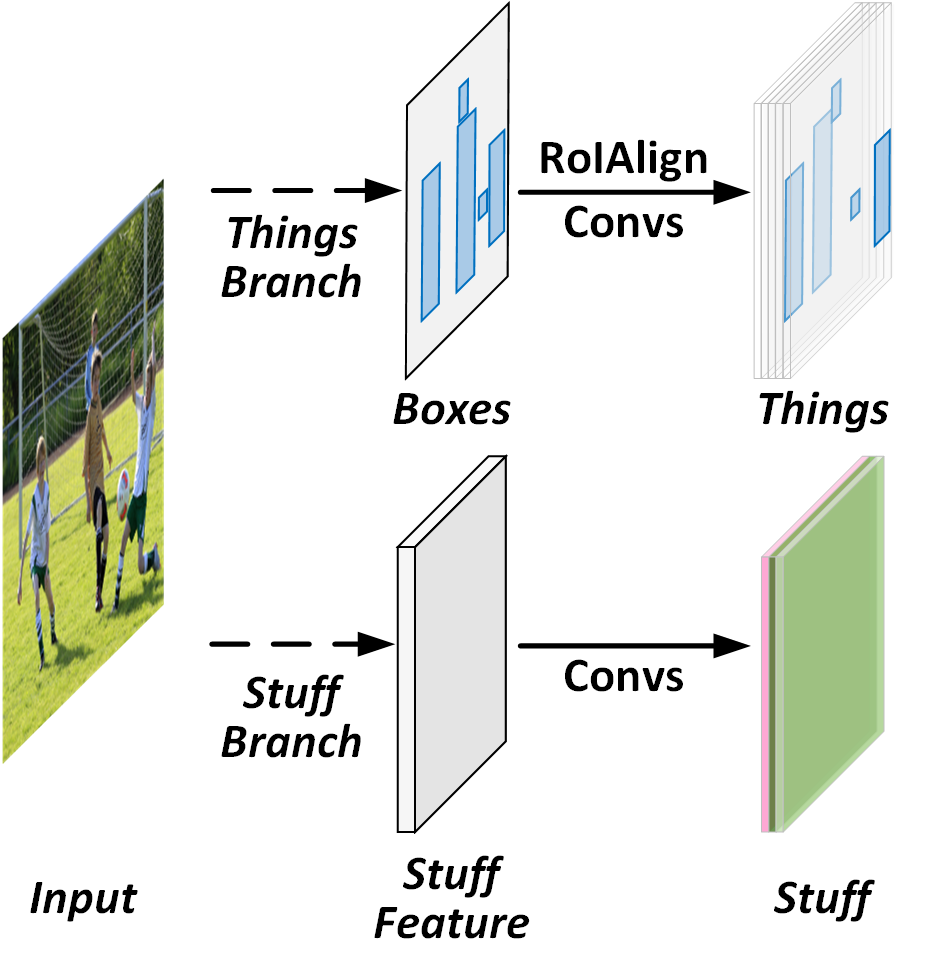}
\label{fig:intro_trad}
}
\quad
\subfigure[Unified representation]{
\includegraphics[width=0.45\linewidth]{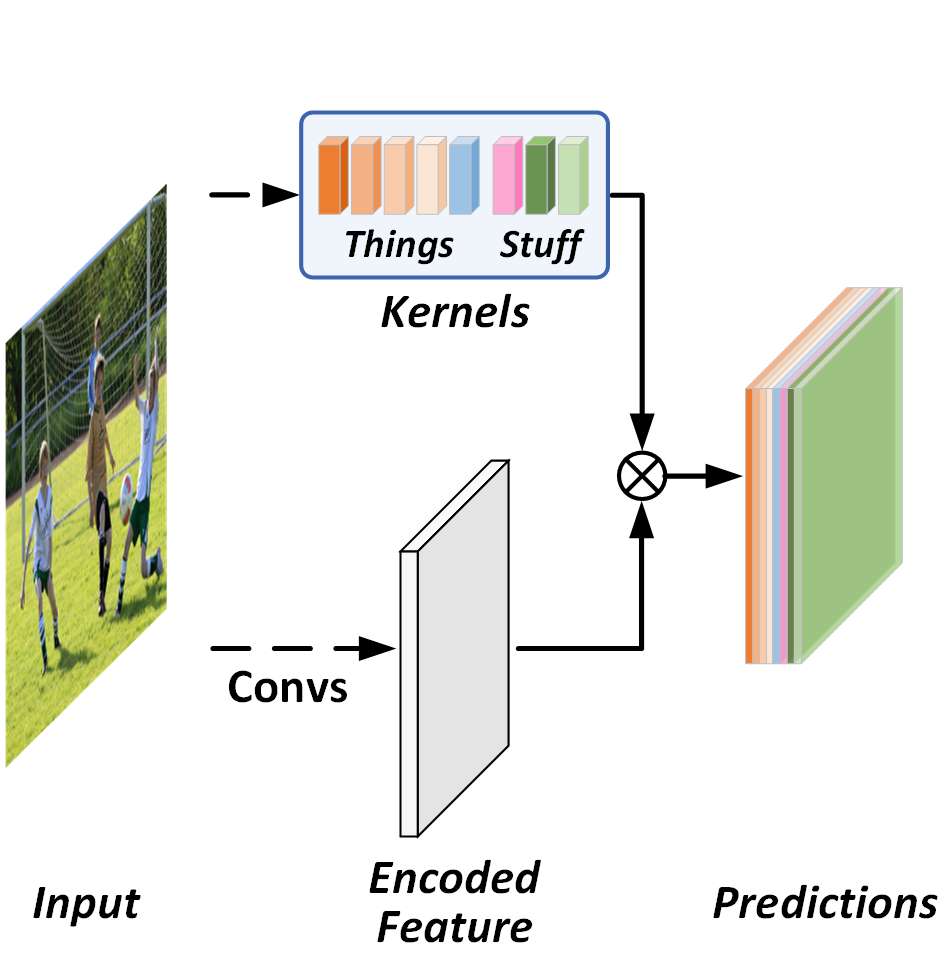}
\label{fig:intro_ours}
}
\caption{Compared with traditional methods, which often utilize separate branches to handle {\em things} and {\em stuff} in~\ref{fig:intro_trad}, the proposed Panoptic FCN~\ref{fig:intro_ours} represents {\em things} and {\em stuff} uniformly with generated kernels. Here, an example with box-based stream for {\em things} is given in~\ref{fig:intro_trad}. The shared backbone is omitted for concision.}
\end{figure}

Panoptic segmentation, aiming to assign each pixel with a semantic label and unique identity, is regarded as a challenging task. In panoptic segmentation~\cite{kirillov2019panoptic}, countable and uncountable instances ({\em i.e.,} things and stuff) are expected to be represented and resolved in a unified workflow. One main difficulty impeding unified representation comes from conflicting properties requested by things and stuff. Specifically, to distinguish among various identities, countable things usually rely on {\em instance-aware} features, which vary with objects. In contrast, uncountable stuff would prefer {\em semantically consistent} characters, which ensures consistent predictions for pixels with the same semantic meaning.
An example is given in Fig.~\ref{fig:intro}, where embedding of {\em individuals} should be diverse for inter-class variations, while characters of {\em grass} should be similar for intra-class consistency.

For conflict at feature level, specific modules are usually tailored for things and stuff separately, as presented in Fig.~\ref{fig:intro_trad}. In particular, {\em instance-aware} demand of things is satisfied mainly from two streams, namely box-based~\cite{kirillov2019panopticfpn, xiong2019upsnet, li2019attention} and box-free~\cite{yang2019deeperlab, gao2019ssap, cheng2020panoptic} methods. Meanwhile, the {\em semantic-consistency} of stuff is met in a pixel-by-pixel manner~\cite{long2015fully}, where similar semantic features would bring identical predictions. A classic case is Panoptic FPN~\cite{kirillov2019panopticfpn}, which utilizes Mask R-CNN~\cite{he2017mask} and FCN~\cite{long2015fully} in separated branches to respectively classify things and stuff, similar to that of Fig.~\ref{fig:intro_trad}. Although attempt~\cite{yang2019deeperlab, gao2019ssap, cheng2020panoptic} has been made to predict things without boxes, extra predictions ({\em e.g.,} affinities~\cite{gao2019ssap}, and offsets~\cite{yang2019deeperlab}) together with post-process procedures are still needed to distinguish among instances, which slow down the whole system and hinder it from being fully convolutional. Consequently, a unified representation is required to bridge this gap. 

In this paper, we propose a fully convolutional framework for unified representation, called {\em Panoptic FCN}. In particular, Panoptic FCN encodes each instance into a specific kernel and generates the prediction by convolutions directly. Thus, both things and stuff can be predicted together with a same resolution. In this way, {\em instance-aware} and {\em semantically consistent} properties for things and stuff can be respectively satisfied in a unified workflow, which is briefly illustrated in Fig.~\ref{fig:intro_ours}. To sum up, the key idea of Panoptic FCN is to {\em represent and predict things and stuff uniformly with generated kernels in a fully convolutional pipeline}. 

To this end, {\em kernel generator} and {\em feature encoder} are respectively designed for {\em kernel weights generation} and {\em shared feature encoding}. Specifically, in kernel generator, we draw inspirations from point-based object detectors~\cite{law2018cornernet, zhou2019objects} and utilize the position head to locate as well as classify foreground objects and background stuff by {\em object centers} and {\em stuff regions}, respectively. Then, we select kernel weights~\cite{jia2016dynamic} with the same positions from the kernel head to represent corresponding instances. For the {\em instance-awareness} and {\em semantic-consistency} described above, a kernel-level operation, called {\em kernel fusion}, is further proposed, which merges kernel weights that are predicted to have the same identity or semantic category. With a naive feature encoder, which preserves the high-resolution feature with details, each prediction of things and stuff can be produced by convolving with generated kernels directly.

In general, the proposed method can be distinguished from two aspects. Firstly, different from previous work for {\em things} generation~\cite{he2017mask,chen2019tensormask,wang2019solo}, which outputs dense predictions and then utilizes NMS for overlaps removal, the deigned framework generates {\em instance-aware} kernels and produces each specific instance directly. Moreover, compared with traditional FCN-based methods for {\em stuff} prediction~\cite{zhao2017pyramid,chen2018encoder,fu2019dual}, which select the most likely category in a pixel-by-pixel manner, our approach aggregates global context into {\em semantically consistent} kernels and presents results of existing semantic classes in a whole-instance manner.

The overall approach, named {\em Panoptic FCN}, can be easily instantiated for panoptic segmentation, which will be fully elaborated in Sec.~\ref{sec:methods}. To demonstrate its superiority, we give extensive ablation studies in Sec.~\ref{sec:ablation}. Furthermore, experimental results are reported on COCO~\cite{lin2014microsoft}, Cityscapes~\cite{cordts2016cityscapes}, and Mapillary Vistas~\cite{neuhold2017mapillary} datasets. Without bells-and-whistles, Panoptic FCN outperforms previous methods with efficiency, and respectively attains {\bf 44.3}\% PQ and {\bf 47.5}\% PQ on COCO {\em val} and {\em test-dev} set. Meanwhile, it surpasses all similar {\em box-free} methods by large margins and achieves leading performance on Cityscapes and Mapillary Vistas {\em val} set with {\bf 61.4}\% PQ and {\bf 36.9}\% PQ, respectively.
\section{Related Work}

\noindent
\textbf{Panoptic segmentation.}
Traditional approaches mainly conduct segmentation for things and stuff separately. The benchmark for panoptic segmentation~\cite{kirillov2019panoptic} directly combines predictions of things and stuff from different models, causing heavy computational overhead. To solve this problem, methods have been proposed by dealing with things and stuff in one model but in separate branches, including Panoptic FPN~\cite{kirillov2019panopticfpn}, AUNet~\cite{li2019attention}, and UPSNet~\cite{xiong2019upsnet}. 
From the view of instance representation, previous work mainly formats things and stuff from different perspectives. Foreground things are usually separated and represented with boxes~\cite{kirillov2019panopticfpn,yang2019sognet,chen2020banet,li2020unifying} or aggregated according to center offsets~\cite{yang2019deeperlab}, while background stuff is often predicted with a parallel FCN~\cite{long2015fully} branch. Although methods of~\cite{li2018weakly,gao2019ssap} represent things and stuff uniformly, the inherent ambiguity cannot be resolved well merely with the pixel-level affinity, which yields the performance drop in complex scenarios. In contrast, the proposed Panoptic FCN represents things and stuff in a uniform and fully convolutional framework with decent performance and efficiency.

\vspace{0.5em}
\noindent
\textbf{Instance segmentation.}
Instance segmentation aims to discriminate objects in the pixel level, which is a finer representation compared with detected boxes. For {\em instance-awareness}, previous works can be roughly divided into two streams, {\em i.e.,} box-based methods and box-free approaches. Box-based methods usually utilize detected boxes to locate or separate objects~\cite{he2017mask,liu2018path,bolya2019yolact,lee2020centermask,qi2020pointins}. Meanwhile, box-free approaches are designed to generate instances without assistance of object boxes~\cite{gao2019ssap,chen2019tensormask,wang2019solo,wang2020solov2}. Recently, AdaptIS~\cite{sofiiuk2019adaptis} and CondInst~\cite{tian2020conditional} are proposed to utilize point-proposal for instance segmentation. However, the instance aggregation or object-level removal is still needed for results. In this paper, we represent objects in a box-free pipeline, which generates the kernel for each object and produces results by convolving the detail-rich feature directly, with no need for object-level duplicates removal~\cite{hu2018relation,qi2018sequential}.

\vspace{0.5em}
\noindent
\textbf{Semantic segmentation.}
Semantic segmentation assigns each pixel with a semantic category, without considering diverse object identities. In recent years, rapid progress has been made on top of FCN~\cite{long2015fully}. Due to the {\em semantically consistent} property, several attempts have been made to capture contextual cues from wider perception fields~\cite{zhao2017pyramid,chen2017rethinking,chen2018encoder} or establish pixel-wise relationship for long-range dependencies~\cite{zhao2018psanet,huang2019ccnet,song2019learnable}. There is also work to design network architectures for semantic segmentation automatically~\cite{liu2019auto,li2020learning}, which is beyond the scope of this paper. Our proposed Panoptic FCN adopts a similar method to represent things and stuff, which aggregates global context into a specific kernel to predict corresponding semantic category.

\begin{figure*}[!tp]
\centering
\includegraphics[width=0.9\linewidth]{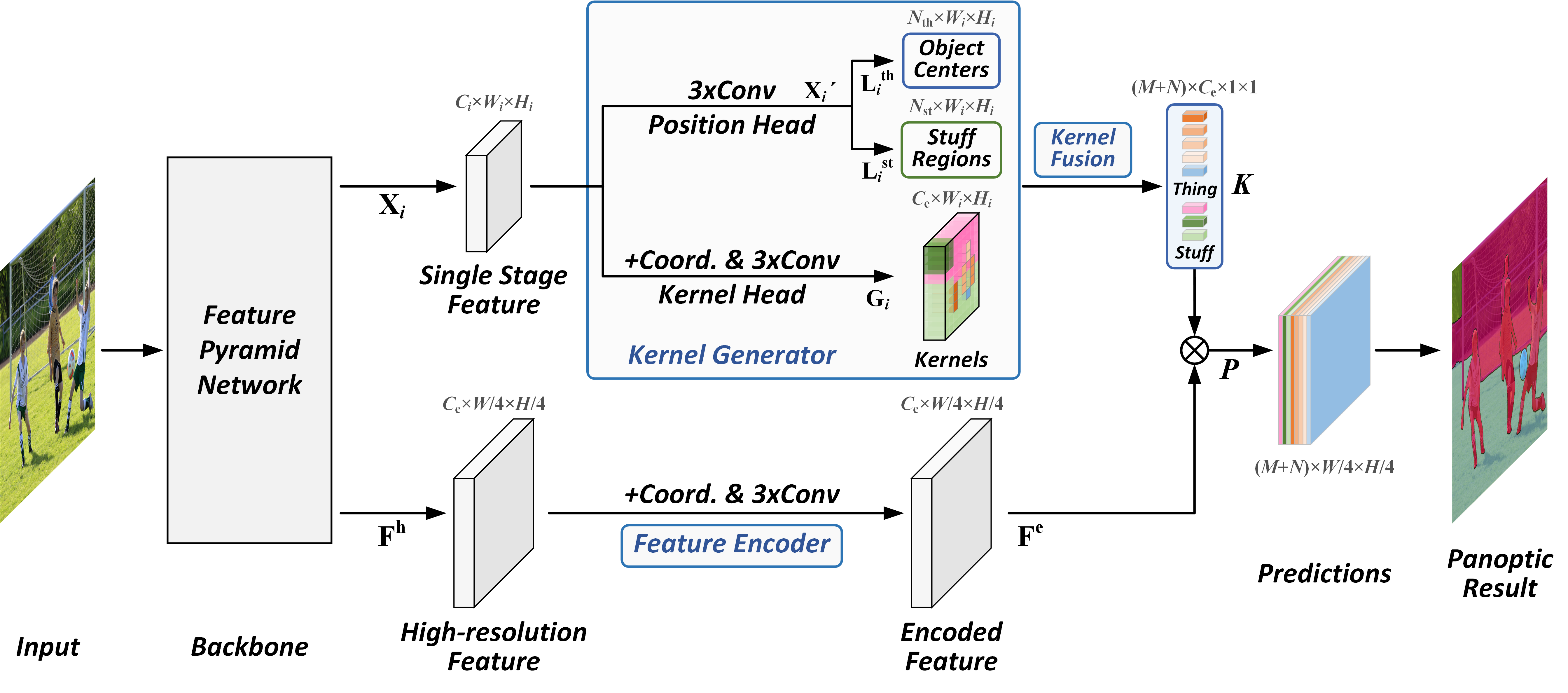} 
\caption{{\bf The framework of Panoptic FCN}. The proposed framework mainly contains {\em three} components, namely {\em kernel generator}, {\em kernel fusion}, and {\em feature encoder}. In {\em kernel generator}, position head is designed to locate and classify object centers along with stuff regions; kernel head in each stage is used to generate kernel weights for both things and stuff. Then, {\em kernel fusion} is utilized to merge kernel weights with the same identity from different stages. And {\em feature encoder} is adopted to encode the high-resolution feature with details. With the generated kernel weight for each instance, both things and stuff can be predicted with a simple convolution directly. Best viewed in color.}
\label{fig:main}
\end{figure*}

\section{Panoptic FCN} \label{sec:methods}
Panoptic FCN is conceptually simple: {\em kernel generator} is introduced to generate kernel weights for things and stuff with different categories; {\em kernel fusion} is designed to merge kernel weights with the same identity from multiple stages; and {\em feature encoder} is utilized to encode the high-resolution feature. In this section, we elaborate on the above components as well as the training and inference scheme. 

\subsection{Kernel Generator}
Given a single stage feature $\mathbf{X}_i$ from the $i$-th stage in FPN~\cite{lin2017feature}, the proposed kernel generator aims at generating the kernel weight map $\mathbf{G}_i$ with positions for things $\mathbf{L}_i^{\mathrm{th}}$ and stuff $\mathbf{L}_i^{\mathrm{st}}$, as depicted in Fig.~\ref{fig:main}. 
To this end, {\em position head} is utilized for instance localization and classification, while {\em kernel head} is designed for kernel weight generation.

\vspace{0.5em}
\noindent
\textbf{Position head.} ~\label{sec:position_head}
With the input $\mathbf{X}_{i}\in \mathbb{R}^{C_i\times W_i\times H_i}$, we simply adopt stacks of convolutions to encode the feature map and generate $\mathbf{X}'_{i}$, as presented in Fig.~\ref{fig:main}. Then we need to locate and classify each instance from the shared feature map $\mathbf{X}'_{i}$. However, according to the definition~\cite{kirillov2019panoptic}, things can be distinguished by object centers, while stuff is uncountable. Thus, we adopt {\em object centers} and {\em stuff regions} to respectively represent position of each individual and stuff category. It means background regions with the same semantic meaning are viewed as one instance. In particular, object map $\mathbf{L}_i^{\mathrm{th}}\in \mathbb{R}^{N_{\mathrm{th}}\times W_i\times H_i}$ and stuff map $\mathbf{L}_i^{\mathrm{st}}\in \mathbb{R}^{N_{\mathrm{st}} \times W_i\times H_i}$ can be generated by convolutions directly with the shared feature map $\mathbf{X}'_{i}$, where $N_{\mathrm{th}}$ and $N_{\mathrm{st}}$ denote the number of semantic category for things and stuff, respectively. 

To better optimize $\mathbf{L}_i^{\mathrm{th}}$ and $\mathbf{L}_i^{\mathrm{st}}$, different strategies are adopted to generate the ground truth. For the $k$-th object in class $c$, we split positive key-points onto the $c$-th channel of the heatmap $\mathbf{Y}_i^{\mathrm{th}}\in\left[0,1\right]^{N_{\mathrm{th}}\times W_i\times H_i}$ with Gaussian kernel, similar to that in~\cite{law2018cornernet, zhou2019objects}. 
With respect to stuff, we produce the ground truth~$\mathbf{Y}_i^{\mathrm{st}}\in\left[0,1\right]^{N_{\mathrm{st}}\times W_i\times H_i}$ by bilinear interpolating the one-hot semantic label to corresponding sizes. Hence, the position head can be optimized with  ${\mathcal L}_{\mathrm{pos}}^{\mathrm{th}}$ and ${\mathcal L}_{\mathrm{pos}}^{\mathrm{st}}$ for object centers and stuff regions, respectively.

\begin{equation} ~\label{equ:loc}
\begin{aligned}
{\mathcal L}_{\mathrm{pos}}^{\mathrm{th}}=&\sum_i\mathrm{FL}(\mathbf{L}_i^{\mathrm{th}},\mathbf{Y}_i^{\mathrm{th}})/N_{\mathrm{th}}, \\
{\mathcal L}_{\mathrm{pos}}^{\mathrm{st}}=&\sum_i\mathrm{FL}(\mathbf{L}_i^{\mathrm{st}},\mathbf{Y}_i^{\mathrm{st}})/{W_i H_i}, \\
{\mathcal L}_{\mathrm{pos}} =& {\mathcal L}_{\mathrm{pos}}^{\mathrm{th}} + {\mathcal L}_{\mathrm{pos}}^{\mathrm{st}},
\end{aligned}
\end{equation}

where $\mathrm{FL}(\cdot,\cdot)$ represents the Focal Loss~\cite{lin2017focal} for optimization.
For inference, $D^{\mathrm{th}}_i=\left\{(x,y):\mathbbm{1}(\mathbf{L}^{\mathrm{th}}_{i,c,x,y})=1\right\}$ and $D^{\mathrm{st}}_i=\left\{(x,y):\mathbbm{1}(\mathbf{L}^{\mathrm{st}}_{i,c,x,y})=1\right\}$ are selected to respectively represent the existence of object centers and stuff regions in corresponding positions with predicted categories $O_i$. This process will be further explained in Sec.~\ref{sec:train_infer}.

\vspace{0.5em}
\noindent
\textbf{Kernel head.}
In kernel head, we first capture spatial cues by directly concatenating relative coordinates to the feature $\mathbf{X}_{i}$, which is similar with that in CoordConv~\cite{liu2018intriguing}. With the concatenated feature map $\mathbf{X}''_{i}\in \mathbb{R}^{(C_i+2)\times W_i\times H_i}$, stacks of convolutions are adopted to generate the kernel weight map $\mathbf{G}_{i}\in \mathbb{R}^{C_\mathrm{e}\times W_i\times H_i}$, as presented in Fig.~\ref{fig:main}. Given predictions $D^{\mathrm{th}}_i$ and $D^{\mathrm{st}}_i$ from the position head, kernel weights with the same coordinates in $\mathbf{G}_{i}$ are chosen to represent corresponding instances. For example, assuming candidate $(x_{c}, y_{c})\in D^{\mathrm{th}}_{i}$, kernel weight $\mathbf{G}_{i, :, x_{c}, y_{c}}\in \mathbb{R}^{C_\mathrm{e}\times1\times1}$ is selected to generate the result with predicted category $c$. The same is true for $D^{\mathrm{st}}_i$. We represent the selected kernel weights in $i$-th stage for things and stuff as $G^{\mathrm{th}}_i$ and $G^{\mathrm{st}}_i$, respectively. Thus, the kernel weight $G^{\mathrm{th}}_i$ and $G^{\mathrm{st}}_i$ together with predicted categories $O_i$ in the $i$-th stage can be produced with the proposed kernel generator.

\subsection{Kernel Fusion} ~\label{sec:kernel_fusion}
Previous work~\cite{ren2015faster,he2017mask,wang2019solo} utilized NMS to remove duplicate boxes or instances in the post-processing stage. Different from them, the designed kernel fusion operation merges repetitive kernel weights from multiple FPN stages before final instance generation, which guarantees {\em instance-awareness} and {\em semantic-consistency} for things and stuff, respectively. In particular, given aggregated kernel weights $G^{\mathrm{th}}$ and $G^{\mathrm{st}}$ from all the stages, the $j$-th kernel weight $K_{j}\in \mathbb{R}^{C_\mathrm{e}\times1\times1}$ is achieved by
\begin{equation}
K_{j}=\mathrm{AvgCluster}(G_{j}'),
\end{equation}
where $\mathrm{AvgCluster}$ denotes average-clustering operation, and the candidate set $G_{j}'=\{G_{m}:\mathrm{ID}(G_{m})=\mathrm{ID}(G_{j})\}$ includes all the kernel weights, which are predicted to have the same identity $\mathrm{ID}$ with $G_{j}$. For object centers, kernel weight $G^{\mathrm{th}}_{m}$ is viewed as identical with $G^{\mathrm{th}}_{j}$ if the {\em cosine similarity} between them surpasses a given threshold $thres$, which will be further investigated in Table~\ref{tab:abla_similar}. For stuff regions, all kernel weights in $G^{\mathrm{st}}$, which share a same category with $G^{\mathrm{st}}_{j}$, are marked as one identity $\mathrm{ID}$. 

With the proposed approach, each kernel weight $K^{\mathrm{th}}_j$ in $K^{\mathrm{th}}=\{K^{\mathrm{th}}_1, ..., K^{\mathrm{th}}_m\}\in \mathbb{R}^{M\times C_\mathrm{e}\times 1 \times 1}$ can be viewed as an embedding for single object, where the total number of objects is $M$. Therefore, kernels with the same identity are merged as a single embedding for things, and each kernel in $K^{\mathrm{th}}$ represents an individual object, which satisfies the {\em instance-awareness} for things. Meanwhile, kernel weight $K^{\mathrm{st}}_j$ in $K^{\mathrm{st}}=\{K^{\mathrm{st}}_1, ..., K^{\mathrm{st}}_n\} \in \mathbb{R}^{N\times C_\mathrm{e}\times 1 \times 1}$ represents the embedding for all $j$-th class pixels, where the existing number of stuff is $N$. With this method, kernels with the same semantic category are fused into a single embedding, which guarantees the {\em semantic-consistency} for stuff. Thus, both properties requested by things and stuff can be fulfilled with the proposed kernel fusion operation.

\subsection{Feature Encoder} ~\label{sec:feature_encoder}
To preserve details for instance representation, high-resolution feature $\mathbf{F}^{\mathrm{h}}\in\mathbb{R}^{C_\mathrm{e}\times W/4\times H/4}$ is utilized for feature encoding. Feature $\mathbf{F}^{\mathrm{h}}$ can be generated from FPN in several ways, {\em e.g.,} P2 stage feature, summed features from all stages, and features from semantic FPN~\cite{kirillov2019panopticfpn}. These methods are compared in Table~\ref{tab:abla_encoder}. Given the feature $\mathbf{F}^{\mathrm{h}}$, a similar strategy with that in kernel head is applied to encode positional cues and generate the encoded feature $\mathbf{F}^{\mathrm{e}}\in\mathbb{R}^{C_\mathrm{e}\times W/4\times H/4}$, as depicted in Fig.~\ref{fig:main}. Thus, given $M$ and $N$ kernel weights for things $K^{\mathrm{th}}$ and stuff $K^{\mathrm{st}}$ from the kernel fusion, each instance is produced by $\mathbf{P}_j=K_j\otimes \mathbf{F}^{\mathrm{e}}$.

Here, $\mathbf{P}_j$ denotes the $j$-th prediction, and $\otimes$ indicates the convolutional operation. That means $M+N$ kernel weights generate $M+N$ instance predictions with resolution $W/4\times H/4$ for the whole image. Consequently, the panoptic result can be produced with a simple process~\cite{kirillov2019panopticfpn}.

\subsection{Training and Inference} \label{sec:train_infer}
\noindent
\textbf{Training scheme.} In the training stage, the central point in each object and all the points in stuff regions are utilized to generate kernel weights for things and stuff, respectively. Here, Dice Loss~\cite{milletari2016v} is adopted to optimize the predicted segmentation, which is formulated as
\begin{equation} ~\label{equ:dice_raw}
{\mathcal L}_{\mathrm{seg}}=\sum_j\mathrm{Dice}(\mathbf{P}_j,\mathbf{Y}_j^{\mathrm{seg}})/(M+N), \\
\end{equation}
where $\mathbf{Y}_j^{\mathrm{seg}}$ denotes ground truth for the $j$-th prediction $\mathbf{P}_j$. To further release the potential of  kernel generator, multiple positives inside each object are sampled to represent the instance. In particular, we select $k$ positions with top predicted scores $s$ inside each object in $\mathbf{L}_i^{\mathrm{th}}$, resulting in $k\times M$ kernels as well as instances in total. This will be explored in Table~\ref{tab:abla_dice}. As for stuff regions, the factor $k$ is set to 1, which means all the points in same category are equally treated. Then, we replace the original loss with a weighted version
\begin{equation}
\mathrm{WDice}(\mathbf{P}_j,\mathbf{Y}_j^{\mathrm{seg}})=\sum_{k}w_k\mathrm{Dice}(\mathbf{P}_{j,k},\mathbf{Y}_{j}^{\mathrm{seg}}), \\
\end{equation}
where $w_k$ denotes the $k$-th weighted score with $w_k=s_k/\sum_{i}s_i$. According to Eqs.~\eqref{equ:loc} and~\eqref{equ:dice_raw}, optimized target ${\mathcal L}$ is defined with the weighted Dice Loss ${\mathcal L}_{\mathrm{seg}}$ as
\begin{equation} ~\label{equ:dice_weighted}
{\mathcal L}_{\mathrm{seg}}=\sum_j\mathrm{WDice}(\mathbf{P}_j,\mathbf{Y}_j^{\mathrm{seg}})/(M+N), \\
\end{equation}
\begin{equation}
{\mathcal L}=\lambda_{\mathrm{pos}}{\mathcal L}_{\mathrm{pos}} + \lambda_{\mathrm{seg}}{\mathcal L}_{\mathrm{seg}}.
\end{equation}

\vspace{0.5em}
\noindent
\textbf{Inference scheme.} In the inference stage, Panoptic FCN follows a simple {\em generate-kernel-then-segment} pipeline. Specifically, we first aggregate positions $D^{\mathrm{th}}_i$, $D^{\mathrm{st}}_i$ and corresponding categories $O_i$ from the $i$-th position head, as illustrated in the Sec.~\ref{sec:position_head}. For object centers, we preserve the peak points in $\mathrm{MaxPool}(\mathbf{L}^{\mathrm{th}}_i)$ utilizing a similar method with that in~\cite{zhou2019objects}. Thus, the indicator for things $\mathbbm{1}(\mathbf{L}^{\mathrm{th}}_{i,c,x,y})$ is marked as positive if point $(x,y)$ in the $c$-th channel is preserved as the peak point. Similarly, the indicator for stuff regions $\mathbbm{1}(\mathbf{L}^{\mathrm{st}}_{i,c,x,y})$ is viewed as positive if point $(x,y)$ with category $c$ is kept. With the designed kernel fusion and the feature encoder, the prediction $\mathbf{P}$ can be easily produced. Specifically, we keep the top 100 scoring kernels of objects and all the kernels of stuff after kernel fusion for instance generation. The threshold 0.4 is utilized to convert predicted soft masks to binary results. It should be noted that both the heuristic process or direct $argmax$ could be used to generate non-overlap panoptic results. The $argmax$ could accelerate the inference but bring performance drop (1.4\% PQ). For fair comparison both from speed and accuracy, the heuristic procedure~\cite{kirillov2019panopticfpn} is adopted in experiments.

\section{Experiments}
In this section, we first introduce experimental settings for Panoptic FCN. Then we conduct abundant studies on the COCO~\cite{lin2014microsoft} {\em val} set to reveal the effect of each component. Finally, comparison with previous methods on COCO~\cite{lin2014microsoft}, Cityscapes~\cite{cordts2016cityscapes}, and Mapillary Vistas~\cite{cordts2016cityscapes} dataset is reported.

\subsection{Experimental Setting}
\noindent
\textbf{Architecture.} From the perspective of network architecture, ResNet~\cite{he2016deep} with FPN~\cite{lin2017feature} are utilized for backbone instantiation. P3 to P7 stages in FPN are used to provide single stage feature $\mathbf{X}_{i}$ for the kernel generator that is shared across all stages. Meanwhile, P2 to P5 stages are adopted to generate the high-resolution feature $\mathbf{F}^\mathrm{h}$, which 
will be further investigated in Table~\ref{tab:abla_encoder}. All convolutions in kernel generator are equipped with $\mathrm{Group Norm}$~\cite{wu2018group} and $\mathrm{ReLU}$ activation. Moreover, a naive convolution is adopted at the end of each head in kernel generator for feature projection.

\vspace{0.5em}
\noindent
\textbf{Datasets.} COCO dataset~\cite{lin2014microsoft} is a widely used benchmark, which contains 80 {\em thing} classes and 53 {\em stuff} classes. It involves 118K, 5K, and 20K images for training, validation, and testing, respectively. Cityscapes dataset~\cite{cordts2016cityscapes} consists of 5,000 street-view {\em fine} annotations with size $1024\times2048$, which are divided into 2,975, 500, and 1,525 images for training, validation, and testing, respectively. Mapillary Vistas~\cite{cordts2016cityscapes} is a traffic-related dataset with resolutions ranging from $1024\times768$ to more than $4000\times6000$. It includes 37 {\em thing} classes and 28 {\em stuff} classes with 18K, 2K, and 5K images for training, validation, and testing, respectively. 

\vspace{0.5em}
\noindent
\textbf{Optimization.} Network optimization is conducted using SGD with weight decay $1e^{-4}$ and momentum 0.9. And {\em poly} schedule with power 0.9 is adopted. Experimentally, $\lambda_{\mathrm{pos}}$ is set to a constant 1, and $\lambda_{\mathrm{seg}}$ are respectively set to 3, 4, and 3 for COCO, Cityscapes, and Mapillary Vistas datasets. For COCO, we set initial rate to 0.01 and follow the $1\times$ strategy in $\mathrm{Detectron2}$~\cite{wu2019detectron2} by default. We randomly flip and rescale the shorter edge from 640 to 800 pixels with 90K iterations. Herein, annotated object centers with instance scale range \{(1,64), (32,128), (64,256), (128,512), (256,2048)\} are assigned to P3-P7 stages, respectively. For Cityscapes, we optimize the network for 65K iterations with an initial rate 0.02 and construct each mini-batch with 32 random $512\times1024$ crops from images that are randomly rescaled from 0.5 to 2.0$\times$. For Mapillary Vistas, the network is optimized for 150K iterations with an initial rate 0.02. In each iteration, we randomly resize images from 1024 to 2048 pixels at the shorted side and build 32 crops with the size $1024\times1024$. Due to the variation in scale distribution, we modify the assigning strategy to \{(1,128), (64,256), (128,512), (256,1024), (512,2048)\} for Cityscapes and Mapillary Vistas datasets.

\begin{table}[t!]
\centering
 \caption{Comparison with different settings of the kernel generator on the COCO {\em val} set. {\em deform} and {\em conv num} respectively denote deformable convolutions for position head and number of convolutions in both heads of the kernel generator.}
\resizebox{0.48\textwidth}{16mm}{
\begin{tabular}{ccccccc}
  \toprule
  {\em {deform}} & {\em{conv num}} & PQ & PQ$^\mathrm{th}$ & PQ$^\mathrm{st}$ & AP & mIoU\\
  \midrule
  \xmark & 1 & 38.4 & 43.4 & 31.0 & 28.3 & 39.9 \\
  \xmark & 2 & 38.9 & 44.1 & 31.1 & 28.9 & 40.1 \\
  \xmark & 3 & 39.2 & 44.7 & 31.0 & 29.6 & 40.2 \\
  \xmark & 4 & 39.2 & 44.9 & 30.8 & 29.4 & 39.9 \\
  \midrule
  \cmark & 3 & {\bf 39.9} & {\bf 45.0} & {\bf 32.4} & {\bf 29.9} & {\bf 41.2} \\
  \bottomrule
\end{tabular}
 }
\label{tab:abla_kernel}
\end{table}

\begin{table}[t!]
\centering
 \caption{Comparison with different positional settings on the COCO {\em val} set. {\em coord}$_\mathrm{w}$ and {\em coord}$_\mathrm{f}$ denote combining coordinates for the kernel head, and feature encoder, respectively.}
\begin{tabular}{ccccccc}
  \toprule
  {\em coord}$_\mathrm{w}$ & {\em coord}$_\mathrm{f}$ & PQ & PQ$^\mathrm{th}$ & PQ$^\mathrm{st}$ & AP & mIoU\\
  \midrule
  \xmark & \xmark & 39.9 & 45.0 & 32.4 & 29.9 & 41.2 \\
  \midrule
  \cmark & \xmark & 39.9 & 45.0 & 32.2 & 30.0 & 41.1 \\
  \xmark & \cmark & 40.2 & 45.3 & 32.5 & 30.4 & 41.6 \\
  \cmark & \cmark & {\bf 41.3} & {\bf 46.9} & {\bf 32.9} & {\bf 32.1} & {\bf 41.7} \\
  \bottomrule
\end{tabular}
 \label{tab:abla_position}
\end{table}

\begin{table}[t!]
\centering
 \caption{Comparison with different similarity thresholds of kernel fusion on the COCO {\em val} set. {\em{class-aware}} denotes only merging kernel weights with the same predicted class. And {\em thres} indicates the cosine similarity threshold {\em thres} for kernel fusion in Sec.~\ref{sec:kernel_fusion}.}
\resizebox{0.48\textwidth}{17mm}{
\begin{tabular}{ccccccc}
  \toprule
   {\em{class-aware}} & {\em{thres}} & PQ & PQ$^\mathrm{th}$ & PQ$^\mathrm{st}$ & AP & mIoU\\
  \midrule
  \cmark & 0.80 & 39.7 & 44.3 & 32.9 & 29.9 & 41.7 \\
  \cmark & 0.85 & 40.8 & 46.1 & 32.9 & 31.5 & 41.7 \\
  \cmark & 0.90 & {\bf 41.3} & 46.9 & {\bf 32.9} & {\bf 32.1} & {\bf 41.7} \\
  \cmark & 0.95 & 41.3 & {\bf 47.0} & 32.9 & 31.1 & 41.7 \\
  \cmark & 1.00 & 38.7 & 42.6 & 32.9 & 25.4 & 41.7 \\
  \midrule
  \xmark & 0.90 & 41.2 & 46.7 & 32.9 & 30.9 & 41.7 \\
  \bottomrule
\end{tabular}
 }
 \label{tab:abla_similar}
\end{table}

\begin{table}[t!]
\centering
 \caption{Comparison with different methods of removing repetitive predictions. \em{kernel-fusion} and \em{nms} indicates the proposed kernel-level fusion method and Matrix NMS~\cite{wang2020solov2}, respectively.}
\resizebox{0.48\textwidth}{13mm}{
\begin{tabular}{ccccccc}
\toprule
   {\em{kernel-fusion}} & {\em{nms}} & PQ & PQ$^\mathrm{th}$ & PQ$^\mathrm{st}$ & AP & mIoU\\
   \midrule
   \xmark & \xmark & 38.7 & 42.6 & 32.9 & 25.4 & 41.7 \\
   \xmark & \cmark & 38.7 & 42.6 & 32.9 & 27.8 & 41.7 \\
   \cmark & \xmark & {\bf 41.3} & {\bf 46.9} & {\bf 32.9} & 32.1 & {\bf 41.7} \\
   \cmark & \cmark & 41.3 & 46.9 & 32.8 & {\bf 32.3} & 41.7 \\
  \bottomrule
\end{tabular}
 }
 \label{tab:abla_nms}
\end{table}

\subsection{Component-wise Analysis} \label{sec:ablation}
\noindent
\textbf{Kernel generator.} Kernel generator plays a vital role in Panoptic FCN. Here, we compare several settings inside kernel generator to improve the kernel expressiveness in each stage. As presented in Table~\ref{tab:abla_kernel}, with the number of convolutions in each head increasing, the network performance improves steadily and achieves the peak PQ with 3 stacked $\mathrm{Conv3\times3}$ whose channel number is 256. Similar with~\cite{zhou2019objects}, deformable convolutions~\cite{zhu2019deformable} are adopted in position head to extend the receptive field, which brings further improvement, especially in stuff regions (1.4\% PQ). 

\vspace{0.5em}
\noindent
\textbf{Position embedding.} Due to the {\em instance-aware} property of objects, position embedding is introduced to provide essential cues. In Table~\ref{tab:abla_position}, we compare among several positional settings by attaching relative coordinates~\cite{liu2018intriguing} to different heads. An interesting finding is that the improvement is minor (up to 0.3\% PQ) if coordinates are attached to the kernel head or feature encoder only, but it boosts to 1.4\% PQ when given the positional cues to both heads. It can be attributed to the constructed correspondence in the position between kernel weights and the encoded feature.

\begin{table}[t!]
\centering
 \caption{Comparison with different channel numbers of the feature encoder on the COCO {\em val} set. {\em channel num} represents the channel number $C_\mathrm{e}$ of the feature encoder.}
\begin{tabular}{cccccc}
  \toprule
  {\small \em{channel num}} & PQ & PQ$^\mathrm{th}$ & PQ$^\mathrm{st}$ & AP & mIoU\\
  \midrule
  16 & 39.9 & 45.0 & 32.1 & 30.8 & 41.3 \\
  32 & 40.8 & 46.3 & 32.5 & 31.7 & 41.6 \\
  64 & {\bf 41.3} & 46.9 & {\bf 32.9} & 32.1 & {\bf 41.7} \\
  128 & 41.3 & {\bf 47.0} & 32.6 & {\bf 32.6} & 41.7 \\
  \bottomrule
\end{tabular}
 \label{tab:abla_channel}
\end{table}

\begin{table}[t!]
\centering
 \caption{Comparison with different feature types for the feature encoder on the COCO {\em val} set. {\em feature type} denotes the method to generate high-resolution feature $\mathbf{F}^{\mathrm{h}}$ in Sec.~\ref{sec:feature_encoder}.}
\resizebox{0.48\textwidth}{11mm}{
\begin{tabular}{lccccc}
  \toprule
  {\em{feature type}} & PQ & PQ$^\mathrm{th}$ & PQ$^\mathrm{st}$ & AP & mIoU\\
  \midrule
  FPN-P2 & 40.6 & 46.0 & 32.4 & 31.6 & 41.3 \\
  FPN-Summed & 40.5 & 46.0 & 32.1 & 31.7 & 41.1 \\
  Semantic FPN~\cite{kirillov2019panopticfpn} & {\bf 41.3} & {\bf 46.9} & {\bf 32.9} & {\bf 32.1} & {\bf 41.7} \\
  \bottomrule
\end{tabular}
 }
\label{tab:abla_encoder}
\end{table}

\begin{table}[t!]
\centering
 \caption{Comparison with different settings of weighted dice loss on the COCO {\em val} set. {\em weighted} and $k$ denote weighted dice loss and the number of sampled points in Sec.~\ref{sec:train_infer}, respectively.}
\begin{tabular}{ccccccc}
  \toprule
  {\small \em weighted } & $k$ & PQ & PQ$^\mathrm{th}$ & PQ$^\mathrm{st}$ & AP & mIoU\\
  \midrule
  \xmark & - & 40.2 & 45.5 & 32.4 & 31.0 & 41.3 \\
  \midrule
  \cmark & 1 & 40.0 & 45.1 & 32.4 & 30.9 & 41.4 \\
  \cmark & 3 & 41.0 & 46.4 & 32.7 & 31.6 & 41.4 \\
  \cmark & 5 & 41.0 & 46.5 & 32.9 & 32.1 & 41.7 \\
  \cmark & 7 & {\bf 41.3} & {\bf 46.9} & {\bf 32.9} & {\bf 32.1} & 41.7 \\
  \cmark & 9 & 41.3 & 46.8 & 32.9 & 32.1 & {\bf 41.8} \\
  \bottomrule
\end{tabular}
 \label{tab:abla_dice}
\end{table}

\begin{table}[t!]
\centering
 \caption{Comparison with different training schedules on the COCO {\em val} set. $\mathrm{1\times}$, $\mathrm{2\times}$, and $\mathrm{3\times}$ {\em schedule} denote the $\mathrm{90K}$, $\mathrm{180K}$, and  $\mathrm{270K}$ training iterations in $\mathrm{Detectron2}$~\cite{wu2019detectron2}, respectively.}
\begin{tabular}{cccccc}
  \toprule
  {\small \em{schedule}} & PQ & PQ$^\mathrm{th}$ & PQ$^\mathrm{st}$ & AP & mIoU\\
  \midrule
  $\mathrm{1\times}$ & 41.3 & 46.9 & 32.9 & 32.1 & 41.7 \\
  $\mathrm{2\times}$ & 43.2 & 48.8 & 34.7 & 34.3 & 43.4 \\
  $\mathrm{3\times}$ & {\bf 43.6} & {\bf 49.3} & {\bf 35.0} & {\bf 34.5} & {\bf 43.8} \\
  \bottomrule
\end{tabular}
 \label{tab:abla_training}
\end{table}

\begin{table}[t!]
\centering
 \caption{Comparison with different settings of the feature encoder on the COCO {\em val} set. 
 {\em deform} and {\em channel num} represent deformable convolutions and the channel number $C_\mathrm{e}$, respectively.}
\resizebox{0.48\textwidth}{8.5mm}{
\begin{tabular}{ccccccc}
  \toprule
  {\em{deform}} & {\em{channel num}} & PQ & PQ$^\mathrm{th}$ & PQ$^\mathrm{st}$ & AP & mIoU\\
  \midrule
  \xmark & 64  & 43.6 & 49.3 & 35.0 & 34.5 & 43.8 \\
  \cmark & 256 & {\bf 44.3} & {\bf 50.0} & {\bf 35.6} & {\bf 35.5} & {\bf 44.0} \\
  \bottomrule
\end{tabular}
}
 \label{tab:abla_enhance}
\end{table}

\begin{table}[t!]
\centering
 \caption{Upper-bound analysis on the COCO {\em val} set. {\em{gt position}} and {\em{gt class}} denote utilizing the ground-truth position $G_i$ and class $O_i$ in each position head for kernel generation, respectively.}
\resizebox{0.48\textwidth}{12.9mm}{
\begin{tabular}{ccccccc}
  \toprule
  {\em{gt position}} & {\em{gt class}} & PQ & PQ$^\mathrm{th}$ & PQ$^\mathrm{st}$ & AP & mIoU\\
  \midrule
  \xmark & \xmark & 43.6 & 49.3 & 35.0 & 34.5 & 43.8 \\
  \cmark & \xmark & 49.8 & 52.2 & 46.1 & 38.2 & 54.6 \\
  \cmark & \cmark & {\bf 65.9} & {\bf 64.1} & {\bf 68.7} & {\bf 45.5} & {\bf 86.6} \\
  \midrule
    &  & {\em +22.3} & {\em +14.8} & {\em +33.7} & {\em +11.0} & {\em +42.8}\\
  \bottomrule
\end{tabular}
}
 \label{tab:abla_upperbound}
\end{table}
\begin{figure}[t!]
\centering
\includegraphics[width=0.98\linewidth]{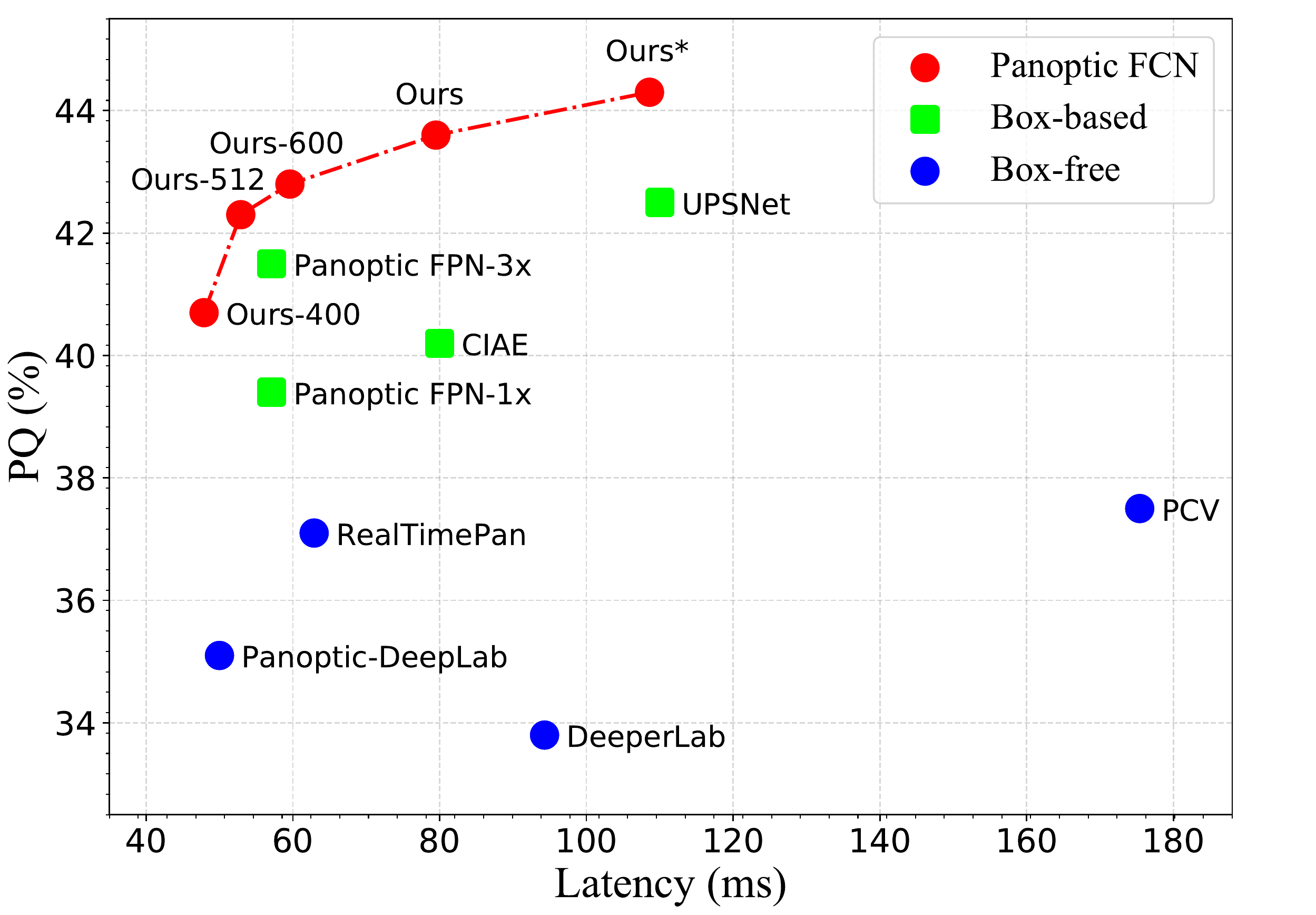} 
\caption{Speed-Accuracy trade-off curve on the COCO {\em val} set. 
All results are compared with Res50 except DeeperLab~\cite{yang2019deeperlab} based on Xception-71~\cite{chollet2017xception}. The latency is measured {\em end-to-end} from single input to panoptic result. Details are given in Table~\ref{tab:coco_val}.
}

\label{fig:speed_acc}
\end{figure}

\vspace{0.5em}
\noindent
\textbf{Kernel fusion.} Kernel fusion is a core operation in the proposed method, which guarantees the required properties for things and stuff, as elaborated in Sec.~\ref{sec:kernel_fusion}. We investigate the fusion type {\em class-aware} and similarity thresholds {\em thres} in Table~\ref{tab:abla_similar}. As shown in the table, the network attains the best performance with {\em thres} 0.90. And the {\em class-agnostic} manner could dismiss some similar instances with different categories, which yields drop in AP. Furthermore, we compare kernel fusion with Matrix NMS~\cite{wang2020solov2} which is utilized for pixel-level removal. As presented in Table~\ref{tab:abla_nms}, the performance saturates with the simple kernel-level fusion method, and extra NMS brings no more gain.

\begin{table*}[t!]
\centering
 \caption{Comparisons with previous methods on the COCO {\em val} set. Panoptic FCN-400, 512, and 600 denotes utilizing smaller input instead of the default setting. All our results are achieved on the same device with single input and no flipping. FPS is measured {\em end-to-end} from single input to panoptic result with an average speed over 1,000 images, which could be further improved with more optimizations. The simple enhanced version is marked with *. The model testing by ourselves according to released codes is denoted as \dag.}
\resizebox{0.98\textwidth}{48mm}{
\begin{tabular}{lcccccccccccc}
  \toprule
  Method & Backbone & PQ & SQ & RQ & PQ$^\mathrm{th}$ & SQ$^\mathrm{th}$ & RQ$^\mathrm{th}$ & PQ$^\mathrm{st}$ & SQ$^\mathrm{st}$ & RQ$^\mathrm{st}$ & Device & FPS\\
  \midrule
  \multicolumn{13}{c}{\small \em box-based} \\
  \midrule
  Panoptic FPN~\cite{kirillov2019panopticfpn} & Res50-FPN & 39.0 & - & - & 45.9 & - & - & 28.7 & - & - & - & - \\
  Panoptic FPN$^\dag$-$\mathrm{1\times}$ & Res50-FPN & 39.4 & 77.8 & 48.3 & 45.9 & 80.9 & 55.3 & 29.6 & 73.3 & 37.7 & V100 & 17.5 \\
  Panoptic FPN$^\dag$-$\mathrm{3\times}$ & Res50-FPN & 41.5 & 79.1 & 50.5 & 48.3 & 82.2 & 57.9 & 31.2 & 74.4 & 39.5 & V100 & 17.5\\
  AUNet~\cite{li2019attention} & Res50-FPN & 39.6 & - & - & 49.1 & - & - & 25.2 & - & - & - & - \\
  CIAE~\cite{gao2020learning} & Res50-FPN & 40.2 & - & - & 45.3 & - & - & 32.3 & - & - & 2080Ti & 12.5 \\
  UPSNet$^\dag$~\cite{xiong2019upsnet} & Res50-FPN & 42.5 & 78.0 & 52.5 & 48.6 & 79.4 & {\bf 59.6} & 33.4 & 75.9 & 41.7 & V100 & 9.1 \\
  Unifying~\cite{li2020unifying} & Res50-FPN & 43.4 & 79.6 & 53.0 & 48.6 & - & - & 35.5 & - & - & - & - \\
  \midrule
  \multicolumn{13}{c}{\small \em box-free} \\
  \midrule
  DeeperLab~\cite{yang2019deeperlab} & Xception-71 & 33.8 & - & - & - & - & - & - & - & - & V100 & 10.6 \\
  Panoptic-DeepLab~\cite{cheng2020panoptic} & Res50 & 35.1 & - & - & - & - & - & - & - & - & V100 & 20.0 \\
  AdaptIS~\cite{sofiiuk2019adaptis} & Res50 & 35.9 & - & - & 40.3 & - & - & 29.3 & - & - & - & - \\
  RealTimePan~\cite{hou2020real} & Res50-FPN & 37.1 & - & - & 41.0 & - & - & 31.3 & - & - & V100 & 15.9 \\
  PCV~\cite{wang2020pixel} & Res50-FPN & 37.5 & 77.7 & 47.2 & 40.0 & 78.4 & 50.0 & 33.7 & 76.5 & 42.9 & 1080Ti & 5.7 \\
  SOLO V2~\cite{wang2020solov2} & Res50-FPN & 42.1 & - & - & 49.6  & - & - & 30.7 & - & -  & - & - \\
  \midrule
  Panoptic FCN-400 & Res50-FPN & 40.7 & 80.5 & 49.3 & 44.9 & 82.0 & 54.0 & 34.3 & 78.1 & 42.1 & V100 & {\bf 20.9} \\
  Panoptic FCN-512 & Res50-FPN & 42.3 & {\bf 80.9} & 51.2 & 47.4 & 82.1 & 56.9 & 34.7 & {\bf 79.1} & 42.7 & V100 & 18.9 \\
  Panoptic FCN-600 & Res50-FPN & 42.8 &	80.6 & 51.6 & 47.9 & 82.6 & 57.2 & 35.1 & 77.4 & 43.1 & V100 & 16.8 \\
  Panoptic FCN & Res50-FPN & 43.6 & 80.6 & 52.6 & 49.3 & 82.6 & 58.9 & 35.0 & 77.6 & 42.9 & V100 & 12.5 \\
  Panoptic FCN$^*$ & Res50-FPN & {\bf 44.3} & 80.7 & {\bf 53.0} & {\bf 50.0} & {\bf 83.4} & 59.3 & {\bf 35.6} & 76.7 & {\bf 43.5} & V100 & 9.2 \\
  \bottomrule
\end{tabular}
 }
 \label{tab:coco_val}
\end{table*}


\vspace{0.5em}
\noindent
\textbf{Feature encoder.} To enhance expressiveness of the encoded feature $\mathbf{F}^\mathrm{e}$, we further explore the {\em channel number} and {\em feature type} used in feature encoder. As illustrated in Table~\ref{tab:abla_channel}, the network achieves 41.3\% PQ with 64 channels, and extra channels contribute little improvement. For efficiency, we set the channel number of feature encoder to 64 by default. As for high-resolution feature generation, three types of methods are further discussed in Table~\ref{tab:abla_encoder}. It is clear that Semantic FPN~\cite{kirillov2019panopticfpn}, which combines features from four stages in FPN, achieves the top performance 41.3\% PQ.

\vspace{0.5em}
\noindent
\textbf{Weighted dice loss.} The designed weighted dice loss aims to release the potential of kernel generator by sampling $k$ positive kernels inside each object. Compared with the original dice loss, which selects a single central point in each object, improvement brought by the weighted dice loss reaches 1.1\% PQ, as presented in Table~\ref{tab:abla_dice}. This is achieved by sampling 7 top-scoring kernels to generate results of each instance, which are optimized together in each step.

\vspace{0.5em}
\noindent
\textbf{Training schedule.} To fully optimize the network, we prolong the training iteration to the $3\times$ training schedule, which is widely adopted in recent one-stage instance-level approaches~\cite{chen2019tensormask,wang2019solo,wang2020solov2}. As shown in Table~\ref{tab:abla_training}, $2\times$ training schedule brings 1.9\% PQ improvements and increasing iterations to $3\times$ schedule contributes extra 0.4\% PQ.

\vspace{0.5em}
\noindent
\textbf{Enhanced version.} We further explore model capacity by combining existing simple enhancements, {\em e.g.,} deformable convolutions and extra channels. As illustrated in Table~\ref{tab:abla_enhance}, the simple reinforcement contributes 0.7\% improvement over the default setting, marked as {\bf Panoptic FCN*}.

\vspace{0.5em}
\noindent
\textbf{Upper-bound analysis.} In Table~\ref{tab:abla_upperbound}, we give analysis to the upper-bound of {\em generate-kernel-then-segment} fashion with Res50-FPN backbone on the COCO {\em val} set. As illustrated in the table, given ground truth positions of object centers $\mathbf{L}_i^{\mathrm{th}}$ and stuff regions $\mathbf{L}_i^{\mathrm{st}}$, the network yields 6.2\% PQ from more precise locations. And it will bring extra boost ({\bf 16.1}\% PQ) to the network if we assign ground truth categories to the position head. Compared with the baseline method, there still remains huge potential to be explored ({\bf 22.3}\% PQ in total), especially for stuff regions which could have even up to {\bf 33.7}\% PQ and {\bf 42.8}\% mIoU gains.

\vspace{0.5em}
\noindent
\textbf{Speed-accuracy.} To verify the network efficiency, we plot the {\em end-to-end} speed-accuracy trade-off curve on the COCO {\em val} set. As presented in Fig.~\ref{fig:speed_acc}, the proposed Panoptic FCN surpasses all previous {\em box-free} models by large margins on both performance and efficiency. Even compared with the well-optimized Panoptic FPN~\cite{kirillov2019panopticfpn} from $\mathrm{Detectron2}$~\cite{wu2019detectron2}, our approach still attains a better speed-accuracy balance with different image scales. Details about these data points are included in Table~\ref{tab:coco_val}.

\begin{table}[th!]
\centering
 \caption{Experiments on the COCO {\em test-dev} set. All our results are achieved with single scale input and no flipping. The simple enhanced version and {\em val} set for training are marked with * and \ddag.}
\resizebox{0.48\textwidth}{39mm}{
\begin{tabular}{lcccc}
  \toprule
  Method & Backbone & PQ & PQ$^\mathrm{th}$ & PQ$^\mathrm{st}$\\
  \midrule
  \multicolumn{5}{c}{\small \em box-based} \\
  \midrule
  Panoptic FPN~\cite{kirillov2019panopticfpn}  & Res101-FPN & 40.9 & 48.3 & 29.7 \\
  CIAE~\cite{gao2020learning} & DCN101-FPN & 44.5 & 49.7 & 36.8 \\
  AUNet~\cite{li2019attention} & ResNeXt152-FPN & 46.5 & {\bf 55.8} & 32.5 \\
  UPSNet~\cite{xiong2019upsnet} & DCN101-FPN & 46.6 & 53.2 & 36.7 \\
  Unifying$^{\ddag}$~\cite{li2020unifying} & DCN101-FPN & 47.2 & 53.5 & 37.7 \\
  BANet~\cite{chen2020banet} & DCN101-FPN & 47.3 & 54.9 & 35.9 \\
  \midrule
  \multicolumn{5}{c}{\small \em box-free} \\
  \midrule
  DeeperLab~\cite{yang2019deeperlab} & Xception-71 & 34.3 & 37.5 & 29.6 \\
  SSAP~\cite{gao2019ssap} & Res101-FPN & 36.9 & 40.1 & 32.0 \\
  PCV~\cite{wang2020pixel} & Res50-FPN & 37.7 & 40.7 & 33.1 \\
  Panoptic-DeepLab~\cite{cheng2020panoptic} & Xception-71 & 39.7 & 43.9 & 33.2 \\
  AdaptIS~\cite{sofiiuk2019adaptis} & ResNeXt-101 & 42.8 & 53.2 & 36.7 \\
  Axial-DeepLab~\cite{wang2020axial} & Axial-ResNet-L & 43.6 & 48.9 & 35.6 \\
  \midrule
  Panoptic FCN & Res101-FPN & 45.5 & 51.4 & 36.4 \\
  Panoptic FCN & DCN101-FPN & 47.0 & 53.0 & 37.8 \\
  Panoptic FCN$^*$ & DCN101-FPN & 47.1 & 53.2 & 37.8 \\
  Panoptic FCN$^{*\ddag}$ & DCN101-FPN & {\bf 47.5} & 53.7 & {\bf 38.2} \\
  \bottomrule
\end{tabular}
 }
 \label{tab:coco_test}
\end{table}
\begin{table}[th!]
\centering

 \caption{Experiments on the Cityscape {\em val} set. All our results are achieved with single scale input and no flipping. The simple enhanced version is marked with *.}
\resizebox{0.48\textwidth}{36mm}{
\begin{tabular}{lcccc}
  \toprule
  Method & Backbone & PQ & PQ$^\mathrm{th}$ & PQ$^\mathrm{st}$\\
  \midrule
  \multicolumn{5}{c}{\small \em box-based} \\
  \midrule
  Panoptic FPN~\cite{kirillov2019panopticfpn} & Res101-FPN & 58.1 & 52.0 & 62.5 \\
  AUNet~\cite{li2019attention} & Res101-FPN & 59.0 & 54.8 & 62.1 \\
  UPSNet~\cite{xiong2019upsnet} & Res50-FPN & 59.3 & 54.6 & 62.7 \\
  SOGNet~\cite{yang2019sognet} & Res50-FPN & 60.0 & {\bf 56.7} & 62.5 \\
  Seamless~\cite{porzi2019seamless} & Res50-FPN & 60.2 & 55.6 & 63.6 \\
  Unifying~\cite{li2020unifying} & Res50-FPN & 61.4 & 54.7 & 66.3 \\
  \midrule
  \multicolumn{5}{c}{\small \em box-free} \\
  \midrule
  PCV~\cite{wang2020pixel} & Res50-FPN & 54.2 & 47.8 & 58.9 \\
  DeeperLab~\cite{yang2019deeperlab} & Xception-71 & 56.5 & - & - \\
  SSAP~\cite{gao2019ssap} & Res50-FPN & 58.4 & 50.6 & - \\
  AdaptIS~\cite{sofiiuk2019adaptis} & Res50 & 59.0 & 55.8 & 61.3 \\
  Panoptic-DeepLab~\cite{cheng2020panoptic} & Res50 & 59.7 & - & - \\
  \midrule
  Panoptic FCN & Res50-FPN & 59.6 & 52.1 & 65.1 \\
  Panoptic FCN$^*$ & Res50-FPN & {\bf 61.4} & 54.8 & {\bf 66.6} \\
  \bottomrule
\end{tabular}

 \label{tab:city_val}
}
\end{table}
\begin{table}[th!]
\centering

 \caption{Experiments on the Mapillary Vistas {\em val} set. All our results are achieved with single scale input and no flipping. The simple enhanced version is marked with *.}
\resizebox{0.48\textwidth}{27mm}{
\begin{tabular}{lcccc}
  \toprule
  Method & Backbone & PQ & PQ$^\mathrm{th}$ & PQ$^\mathrm{st}$\\
  \midrule
  \multicolumn{5}{c}{\small \em box-based} \\
  \midrule
  BGRNet~\cite{wu2020bidirectional} & Res50-FPN & 31.8 & {\bf 34.1} & 27.3 \\
  TASCNet~\cite{li2018learning} & Res50-FPN & 32.6 & 31.1 & 34.4 \\
  Seamless~\cite{porzi2019seamless} & Res50-FPN & 36.2 & 33.6 & 40.0 \\
  \midrule
  \multicolumn{5}{c}{\small \em box-free} \\
  \midrule
  DeeperLab~\cite{yang2019deeperlab} & Xception-71 & 32.0 & - & - \\
  AdaptIS~\cite{sofiiuk2019adaptis} & Res50 & 32.0 & 26.6 & 39.1 \\
  Panoptic-DeepLab~\cite{cheng2020panoptic} & Res50 & 33.3 & - & - \\
  \midrule
  Panoptic FCN & Res50-FPN & 34.8 & 30.6 & 40.5 \\
  Panoptic FCN$^*$ & Res50-FPN & {\bf 36.9} & 32.9 & {\bf 42.3} \\
  \bottomrule
\end{tabular}

 \label{tab:mapillary_val}
}
\end{table}

\subsection{Main Results}
We further conduct experiments on different scenarios, namely COCO dataset for common context, Cityscapes and Mapillary Vistas datasets for traffic-related environments.

\vspace{0.5em}
\noindent
\textbf{COCO.} In Table~\ref{tab:coco_val}, we conduct experiments on COCO {\em val} set. Compared with recent approaches, Panoptic FCN achieves superior performance with efficiency, which surpasses leading {\em box-based}~\cite{li2020unifying} and {\em box-free}~\cite{wang2020pixel} methods over 0.2\% and {\bf 1.5}\% PQ, respectively. With simple enhancement, the gap enlarges to {\bf 0.9}\% and {\bf 2.2}\% PQ. 
Meanwhile, Panoptic FCN outperforms all top-ranking models on COCO {\em test-dev} set, as illustrated in Table~\ref{tab:coco_test}. In particular, the proposed method surpasses the state-of-the-art approach in {\em box-based} stream with 0.2\% PQ and attains {\bf 47.5}\% PQ with single scale inputs. Compared with the similar {\em box-free} fashion, our method improves {\bf 1.9}\% PQ over Axial-DeepLab~\cite{wang2020axial} which adopts stronger backbone.

\vspace{0.5em}
\noindent
\textbf{Cityscapes.} Furthermore, we carry out experiments on Cityscapes {\em val} set in Table~\ref{tab:city_val}. Panoptic FCN exceeds the top {\em box-free} model~\cite{cheng2020panoptic} with {\bf 1.7}\% PQ and attains {\bf 61.4}\% PQ. Even compared with the leading {\em box-based} model~\cite{li2020unifying}, which utilizes Lovasz loss for further optimization, the proposed method still achieves comparable performance. 

\vspace{0.5em}
\noindent
\textbf{Mapillary Vistas.} In Table~\ref{tab:mapillary_val}, we compare with other state-of-the-art models on the large-scale Mapillary Vistas {\em val} set with Res50-FPN backbone. As presented in the table, the proposed Panoptic FCN exceeds previous {\em box-free} methods by a large margin in both things and stuff. Specifically, Panoptic FCN surpasses the leading {\em box-based}~\cite{porzi2019seamless} and {\em box-free}~\cite{cheng2020panoptic} models
with 0.7\% and {\bf 3.6}\% PQ, and attains {\bf 36.9}\% PQ with simple enhancement in the feature encoder.
\section{Conclusion}
We have presented the Panoptic FCN, a conceptually simple yet effective framework for panoptic segmentation. The key difference from prior works lies on that we represent and predict things and stuff in a fully convolutional manner. To this end, {\em kernel generator} and {\em kernel fusion} are proposed to generate the unique kernel weight for each object instance or semantic category. With the high-resolution feature produced by {\em feature encoder}, prediction is achieved by convolutions directly. Meanwhile, {\em instance-awareness} and {\em semantic-consistency} for things and stuff are respectively satisfied with the designed workflow.
\section{Acknowledgment}
This research was partially supported by National Key R\&D Program of China (No. 2017YFA0700800), and Beijing Academy of Artificial Intelligence (BAAI).

\clearpage
\flushend
{\small
\bibliographystyle{ieee_fullname}
\bibliography{egbib}

\begin{thebibliography}{10}\itemsep=-1pt

\bibitem{bolya2019yolact}
Daniel Bolya, Chong Zhou, Fanyi Xiao, and Yong~Jae Lee.
\newblock Yolact: Real-time instance segmentation.
\newblock In {\em ICCV}, 2019.

\bibitem{chen2017rethinking}
Liang-Chieh Chen, George Papandreou, Florian Schroff, and Hartwig Adam.
\newblock Rethinking atrous convolution for semantic image segmentation.
\newblock {\em arXiv:1706.05587}, 2017.

\bibitem{chen2018encoder}
Liang-Chieh Chen, Yukun Zhu, George Papandreou, Florian Schroff, and Hartwig
  Adam.
\newblock Encoder-decoder with atrous separable convolution for semantic image
  segmentation.
\newblock In {\em ECCV}, 2018.

\bibitem{chen2019tensormask}
Xinlei Chen, Ross Girshick, Kaiming He, and Piotr Doll{\'a}r.
\newblock Tensormask: A foundation for dense object segmentation.
\newblock In {\em ICCV}, 2019.

\bibitem{chen2020banet}
Yifeng Chen, Guangchen Lin, Songyuan Li, Omar Bourahla, Yiming Wu, Fangfang
  Wang, Junyi Feng, Mingliang Xu, and Xi Li.
\newblock Banet: Bidirectional aggregation network with occlusion handling for
  panoptic segmentation.
\newblock In {\em CVPR}, 2020.

\bibitem{cheng2020panoptic}
Bowen Cheng, Maxwell~D Collins, Yukun Zhu, Ting Liu, Thomas~S Huang, Hartwig
  Adam, and Liang-Chieh Chen.
\newblock Panoptic-deeplab: A simple, strong, and fast baseline for bottom-up
  panoptic segmentation.
\newblock In {\em CVPR}, 2020.

\bibitem{chollet2017xception}
Fran{\c{c}}ois Chollet.
\newblock Xception: Deep learning with depthwise separable convolutions.
\newblock In {\em CVPR}, 2017.

\bibitem{cordts2016cityscapes}
Marius Cordts, Mohamed Omran, Sebastian Ramos, Timo Rehfeld, Markus Enzweiler,
  Rodrigo Benenson, Uwe Franke, Stefan Roth, and Bernt Schiele.
\newblock The cityscapes dataset for semantic urban scene understanding.
\newblock In {\em CVPR}, 2016.

\bibitem{fu2019dual}
Jun Fu, Jing Liu, Haijie Tian, Yong Li, Yongjun Bao, Zhiwei Fang, and Hanqing
  Lu.
\newblock Dual attention network for scene segmentation.
\newblock In {\em CVPR}, 2019.

\bibitem{gao2019ssap}
Naiyu Gao, Yanhu Shan, Yupei Wang, Xin Zhao, Yinan Yu, Ming Yang, and Kaiqi
  Huang.
\newblock Ssap: Single-shot instance segmentation with affinity pyramid.
\newblock In {\em ICCV}, 2019.

\bibitem{gao2020learning}
Naiyu Gao, Yanhu Shan, Xin Zhao, and Kaiqi Huang.
\newblock Learning category-and instance-aware pixel embedding for fast
  panoptic segmentation.
\newblock {\em arXiv:2009.13342}, 2020.

\bibitem{he2017mask}
Kaiming He, Georgia Gkioxari, Piotr Doll{\'a}r, and Ross Girshick.
\newblock Mask r-cnn.
\newblock In {\em ICCV}, 2017.

\bibitem{he2016deep}
Kaiming He, Xiangyu Zhang, Shaoqing Ren, and Jian Sun.
\newblock Deep residual learning for image recognition.
\newblock In {\em CVPR}, 2016.

\bibitem{hou2020real}
Rui Hou, Jie Li, Arjun Bhargava, Allan Raventos, Vitor Guizilini, Chao Fang,
  Jerome Lynch, and Adrien Gaidon.
\newblock Real-time panoptic segmentation from dense detections.
\newblock In {\em CVPR}, 2020.

\bibitem{hu2018relation}
Han Hu, Jiayuan Gu, Zheng Zhang, Jifeng Dai, and Yichen Wei.
\newblock Relation networks for object detection.
\newblock In {\em CVPR}, 2018.

\bibitem{huang2019ccnet}
Zilong Huang, Xinggang Wang, Lichao Huang, Chang Huang, Yunchao Wei, and Wenyu
  Liu.
\newblock Ccnet: Criss-cross attention for semantic segmentation.
\newblock In {\em ICCV}, 2019.

\bibitem{jia2016dynamic}
Xu Jia, Bert De~Brabandere, Tinne Tuytelaars, and Luc Van~Gool.
\newblock Dynamic filter networks.
\newblock In {\em NeurIPS}, 2016.

\bibitem{kirillov2019panopticfpn}
Alexander Kirillov, Ross Girshick, Kaiming He, and Piotr Doll{\'a}r.
\newblock Panoptic feature pyramid networks.
\newblock In {\em CVPR}, 2019.

\bibitem{kirillov2019panoptic}
Alexander Kirillov, Kaiming He, Ross Girshick, Carsten Rother, and Piotr
  Doll{\'a}r.
\newblock Panoptic segmentation.
\newblock In {\em CVPR}, 2019.

\bibitem{law2018cornernet}
Hei Law and Jia Deng.
\newblock Cornernet: Detecting objects as paired keypoints.
\newblock In {\em ECCV}, 2018.

\bibitem{lee2020centermask}
Youngwan Lee and Jongyoul Park.
\newblock Centermask: Real-time anchor-free instance segmentation.
\newblock In {\em CVPR}, 2020.

\bibitem{li2018learning}
Jie Li, Allan Raventos, Arjun Bhargava, Takaaki Tagawa, and Adrien Gaidon.
\newblock Learning to fuse things and stuff.
\newblock {\em arXiv:1812.01192}, 2018.

\bibitem{li2018weakly}
Qizhu Li, Anurag Arnab, and Philip~HS Torr.
\newblock Weakly-and semi-supervised panoptic segmentation.
\newblock In {\em ECCV}, 2018.

\bibitem{li2020unifying}
Qizhu Li, Xiaojuan Qi, and Philip~HS Torr.
\newblock Unifying training and inference for panoptic segmentation.
\newblock In {\em CVPR}, 2020.

\bibitem{li2019attention}
Yanwei Li, Xinze Chen, Zheng Zhu, Lingxi Xie, Guan Huang, Dalong Du, and
  Xingang Wang.
\newblock Attention-guided unified network for panoptic segmentation.
\newblock In {\em CVPR}, 2019.

\bibitem{li2020learning}
Yanwei Li, Lin Song, Yukang Chen, Zeming Li, Xiangyu Zhang, Xingang Wang, and
  Jian Sun.
\newblock Learning dynamic routing for semantic segmentation.
\newblock In {\em CVPR}, 2020.

\bibitem{lin2017feature}
Tsung-Yi Lin, Piotr Doll{\'a}r, Ross Girshick, Kaiming He, Bharath Hariharan,
  and Serge Belongie.
\newblock Feature pyramid networks for object detection.
\newblock In {\em CVPR}, 2017.

\bibitem{lin2017focal}
Tsung-Yi Lin, Priya Goyal, Ross Girshick, Kaiming He, and Piotr Doll{\'a}r.
\newblock Focal loss for dense object detection.
\newblock In {\em ICCV}, 2017.

\bibitem{lin2014microsoft}
Tsung-Yi Lin, Michael Maire, Serge Belongie, James Hays, Pietro Perona, Deva
  Ramanan, Piotr Doll{\'a}r, and C~Lawrence Zitnick.
\newblock Microsoft coco: Common objects in context.
\newblock In {\em ECCV}, 2014.

\bibitem{liu2019auto}
Chenxi Liu, Liang-Chieh Chen, Florian Schroff, Hartwig Adam, Wei Hua, Alan~L
  Yuille, and Li Fei-Fei.
\newblock Auto-deeplab: Hierarchical neural architecture search for semantic
  image segmentation.
\newblock In {\em CVPR}, 2019.

\bibitem{liu2018intriguing}
Rosanne Liu, Joel Lehman, Piero Molino, Felipe~Petroski Such, Eric Frank, Alex
  Sergeev, and Jason Yosinski.
\newblock An intriguing failing of convolutional neural networks and the
  coordconv solution.
\newblock In {\em NeurIPS}, 2018.

\bibitem{liu2018path}
Shu Liu, Lu Qi, Haifang Qin, Jianping Shi, and Jiaya Jia.
\newblock Path aggregation network for instance segmentation.
\newblock In {\em CVPR}, 2018.

\bibitem{long2015fully}
Jonathan Long, Evan Shelhamer, and Trevor Darrell.
\newblock Fully convolutional networks for semantic segmentation.
\newblock In {\em CVPR}, 2015.

\bibitem{milletari2016v}
Fausto Milletari, Nassir Navab, and Seyed-Ahmad Ahmadi.
\newblock V-net: Fully convolutional neural networks for volumetric medical
  image segmentation.
\newblock In {\em 3DV}, 2016.

\bibitem{neuhold2017mapillary}
Gerhard Neuhold, Tobias Ollmann, Samuel Rota~Bulo, and Peter Kontschieder.
\newblock The mapillary vistas dataset for semantic understanding of street
  scenes.
\newblock In {\em ICCV}, 2017.

\bibitem{porzi2019seamless}
Lorenzo Porzi, Samuel~Rota Bulo, Aleksander Colovic, and Peter Kontschieder.
\newblock Seamless scene segmentation.
\newblock In {\em CVPR}, 2019.

\bibitem{qi2018sequential}
Lu Qi, Shu Liu, Jianping Shi, and Jiaya Jia.
\newblock Sequential context encoding for duplicate removal.
\newblock In {\em NeurIPS}, 2018.

\bibitem{qi2020pointins}
Lu Qi, Xiangyu Zhang, Yingcong Chen, Yukang Chen, Jian Sun, and Jiaya Jia.
\newblock Pointins: Point-based instance segmentation.
\newblock {\em arXiv:2003.06148}, 2020.

\bibitem{ren2015faster}
Shaoqing Ren, Kaiming He, Ross Girshick, and Jian Sun.
\newblock Faster r-cnn: Towards real-time object detection with region proposal
  networks.
\newblock In {\em NeurIPS}, 2015.

\bibitem{sofiiuk2019adaptis}
Konstantin Sofiiuk, Olga Barinova, and Anton Konushin.
\newblock Adaptis: Adaptive instance selection network.
\newblock In {\em ICCV}, 2019.

\bibitem{song2019learnable}
Lin Song, Yanwei Li, Zeming Li, Gang Yu, Hongbin Sun, Jian Sun, and Nanning
  Zheng.
\newblock Learnable tree filter for structure-preserving feature transform.
\newblock In {\em NeurIPS}, 2019.

\bibitem{tian2020conditional}
Zhi Tian, Chunhua Shen, and Hao Chen.
\newblock Conditional convolutions for instance segmentation.
\newblock {\em arXiv preprint arXiv:2003.05664}, 2020.

\bibitem{wang2020pixel}
Haochen Wang, Ruotian Luo, Michael Maire, and Greg Shakhnarovich.
\newblock Pixel consensus voting for panoptic segmentation.
\newblock In {\em CVPR}, 2020.

\bibitem{wang2020axial}
Huiyu Wang, Yukun Zhu, Bradley Green, Hartwig Adam, Alan Yuille, and
  Liang-Chieh Chen.
\newblock Axial-deeplab: Stand-alone axial-attention for panoptic segmentation.
\newblock In {\em ECCV}, 2020.

\bibitem{wang2019solo}
Xinlong Wang, Tao Kong, Chunhua Shen, Yuning Jiang, and Lei Li.
\newblock Solo: Segmenting objects by locations.
\newblock In {\em ECCV}, 2020.

\bibitem{wang2020solov2}
Xinlong Wang, Rufeng Zhang, Tao Kong, Lei Li, and Chunhua Shen.
\newblock Solov2: Dynamic, faster and stronger.
\newblock In {\em NeurIPS}, 2020.

\bibitem{wu2018group}
Yuxin Wu and Kaiming He.
\newblock Group normalization.
\newblock In {\em ECCV}, 2018.

\bibitem{wu2019detectron2}
Yuxin Wu, Alexander Kirillov, Francisco Massa, Wan-Yen Lo, and Ross Girshick.
\newblock Detectron2.
\newblock \url{https://github.com/facebookresearch/detectron2}, 2019.

\bibitem{wu2020bidirectional}
Yangxin Wu, Gengwei Zhang, Yiming Gao, Xiajun Deng, Ke Gong, Xiaodan Liang, and
  Liang Lin.
\newblock Bidirectional graph reasoning network for panoptic segmentation.
\newblock In {\em CVPR}, 2020.

\bibitem{xiong2019upsnet}
Yuwen Xiong, Renjie Liao, Hengshuang Zhao, Rui Hu, Min Bai, Ersin Yumer, and
  Raquel Urtasun.
\newblock Upsnet: A unified panoptic segmentation network.
\newblock In {\em CVPR}, 2019.

\bibitem{yang2019deeperlab}
Tien-Ju Yang, Maxwell~D Collins, Yukun Zhu, Jyh-Jing Hwang, Ting Liu, Xiao
  Zhang, Vivienne Sze, George Papandreou, and Liang-Chieh Chen.
\newblock Deeperlab: Single-shot image parser.
\newblock {\em arXiv:1902.05093}, 2019.

\bibitem{yang2019sognet}
Yibo Yang, Hongyang Li, Xia Li, Qijie Zhao, Jianlong Wu, and Zhouchen Lin.
\newblock Sognet: Scene overlap graph network for panoptic segmentation.
\newblock In {\em AAAI}, 2020.

\bibitem{zhao2017pyramid}
Hengshuang Zhao, Jianping Shi, Xiaojuan Qi, Xiaogang Wang, and Jiaya Jia.
\newblock Pyramid scene parsing network.
\newblock In {\em CVPR}, 2017.

\bibitem{zhao2018psanet}
Hengshuang Zhao, Yi Zhang, Shu Liu, Jianping Shi, Chen Change~Loy, Dahua Lin,
  and Jiaya Jia.
\newblock Psanet: Point-wise spatial attention network for scene parsing.
\newblock In {\em ECCV}, 2018.

\bibitem{zhou2019objects}
Xingyi Zhou, Dequan Wang, and Philipp Kr{\"a}henb{\"u}hl.
\newblock Objects as points.
\newblock {\em arXiv:1904.07850}, 2019.

\bibitem{zhu2019deformable}
Xizhou Zhu, Han Hu, Stephen Lin, and Jifeng Dai.
\newblock Deformable convnets v2: More deformable, better results.
\newblock In {\em CVPR}, 2019.

\end{thebibliography}
}

\clearpage

\appendix

\section{Experimental Details}

Herein, we provide more technical details of the {\em training} and {\em inference} process in the proposed Panoptic FCN.

\subsection{Training Details} 
\noindent
{\bf Position head.}
To better optimize the position head, we have explored different types of center assigning strategy. For object center generation, {\em mass center} is utilized to provide the $k$-th ground-truth coordinate $\widetilde x_k$ and $\widetilde y_k$ of the main paper. As presented in Table~\ref{tab:abla_centertype}, compared with the {\em box center} for ground-truth generation, we find that {\em mass center} brings superior performance and higher robustness, especially in the designed weighted dice loss, which samples 7 top-scoring points inside each object. It could be attributed to that most of mass centers are located within the object area, while it is not the case for box centers. Moreover, the object size-adaptive deviation $\sigma_k$ of the main paper is set to $(2r+1)/3$, where $r$ denotes the gaussian radius of the $k$-th object similar to that in CornerNet~\cite{law2018cornernet}.

\vspace{0.5em}
\noindent
{\bf Weighted dice loss.} For objects, we select $k$ positions whose scores are predicted to be top-$k$ within the object region of $\mathbf{L}_{i,c}^\mathrm{th}$, where $c$ is the annotated class. This procedure utilizes $k$ top-scoring kernels to represent the same object and generates $k$ results of prediction $\mathbf{P}_j$. Thus, we generate the same ground-truth $\mathbf{Y}_j^\mathrm{seg}$ for $k$ results in $\mathbf{P}_j$, which can be optimized with Eq.~\eqref{equ:dice_weighted}. For stuff regions, we merge all positions with the same category in each stage using $\mathrm{AvgCluster}$, which brings a specific kernel. Hence, the factor $k$ can be viewed as 1 for each stuff region.

\begin{table}[!ht]
\centering
 \caption{Comparisons among different settings of center type on the COCO {\em val} set. {\em weighted} and {\em center type} denote weighted dice loss and center type for ground-truth generation, respectively.}
\resizebox{0.48\textwidth}{13.2mm}{
\begin{tabular}{ccccccc}
  \toprule
  {\small \em weighted }& {\small \em center type} & PQ & PQ$^\mathrm{th}$ & PQ$^\mathrm{st}$ & AP & mIoU\\
  \midrule
  \xmark & box & 39.7 & 44.7 & 32.3 & 30.3 & 41.2 \\
  \xmark & mass & 40.2 & 45.5 & 32.4 & 31.0 & 41.3 \\
  \midrule
  \cmark & box & 40.6 & 45.7 & 32.8 & 31.7 & 41.5 \\
  \cmark & mass & {\bf 41.3} & {\bf 46.9} & {\bf 32.9} & {\bf 32.1} & {\bf 41.7} \\
  \bottomrule
\end{tabular}
}
 \label{tab:abla_centertype}
\end{table}

\vspace{0.5em}
\noindent
\subsection{Inference Details.}
Due to the variation of scale distribution among datasets, we simply modify some parameters in the inference stage. In particular, for COCO dataset, we preserve 100 top-scoring kernels for object prediction and utilize threshold 0.4 to convert soft masks to binary results, as illustrated in Sec.~\ref{sec:train_infer} of the main paper. For Cityscapes and Mapillary Vistas datasets, 200 kernels with top predicted scores are kept for object prediction, and cosine threshold 0.95 and mask threshold 0.5 are utilized to fuse repetitive kernels and convert binary masks, respectively. Meanwhile, a similar strategy with that in SOLO~\cite{wang2019solo,wang2020solov2} is adopted to adjust predicted object scores. Furthermore, our codes for training and inference will be released to provide more details.

\section{Qualitative Results}
We further visualize qualitative results of Panoptic FCN on several datasets with {\em common context} and {\em traffic-related scenarios},~{\em i.e.}, COCO, Cityscapes, and Mapillary Vistas.

\begin{figure*}[h!]
\centering
\includegraphics[width=\linewidth]{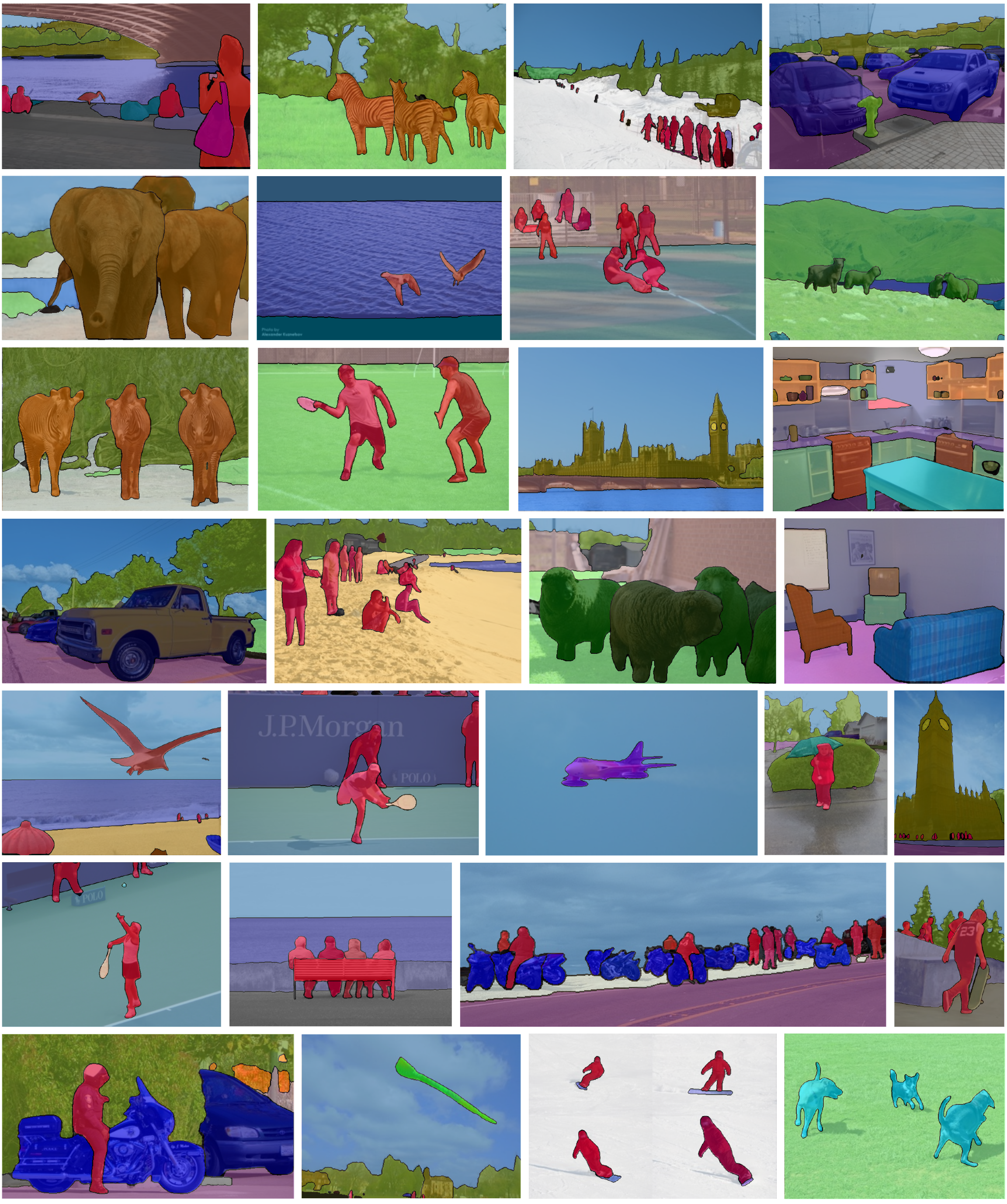} 
\caption{Visualization of panoptic results on the COCO {\em val} set.
}
\label{fig:vis_coco}
\end{figure*}

\vspace{0.5em}
\noindent
{\bf COCO.} As presented in Fig.~\ref{fig:vis_coco}, Panoptic FCN gives detailed characterization to the daily environment. Thanks to the {\em pixel-by-pixel} handling manner and unified representation, details in foreground things and background stuff can be preserved. Moreover, the proposed approach also validates its effectiveness on objects with various scales in Fig.~\ref{fig:vis_coco}.

\begin{figure*}[t!]
\centering
\includegraphics[width=0.85\linewidth]{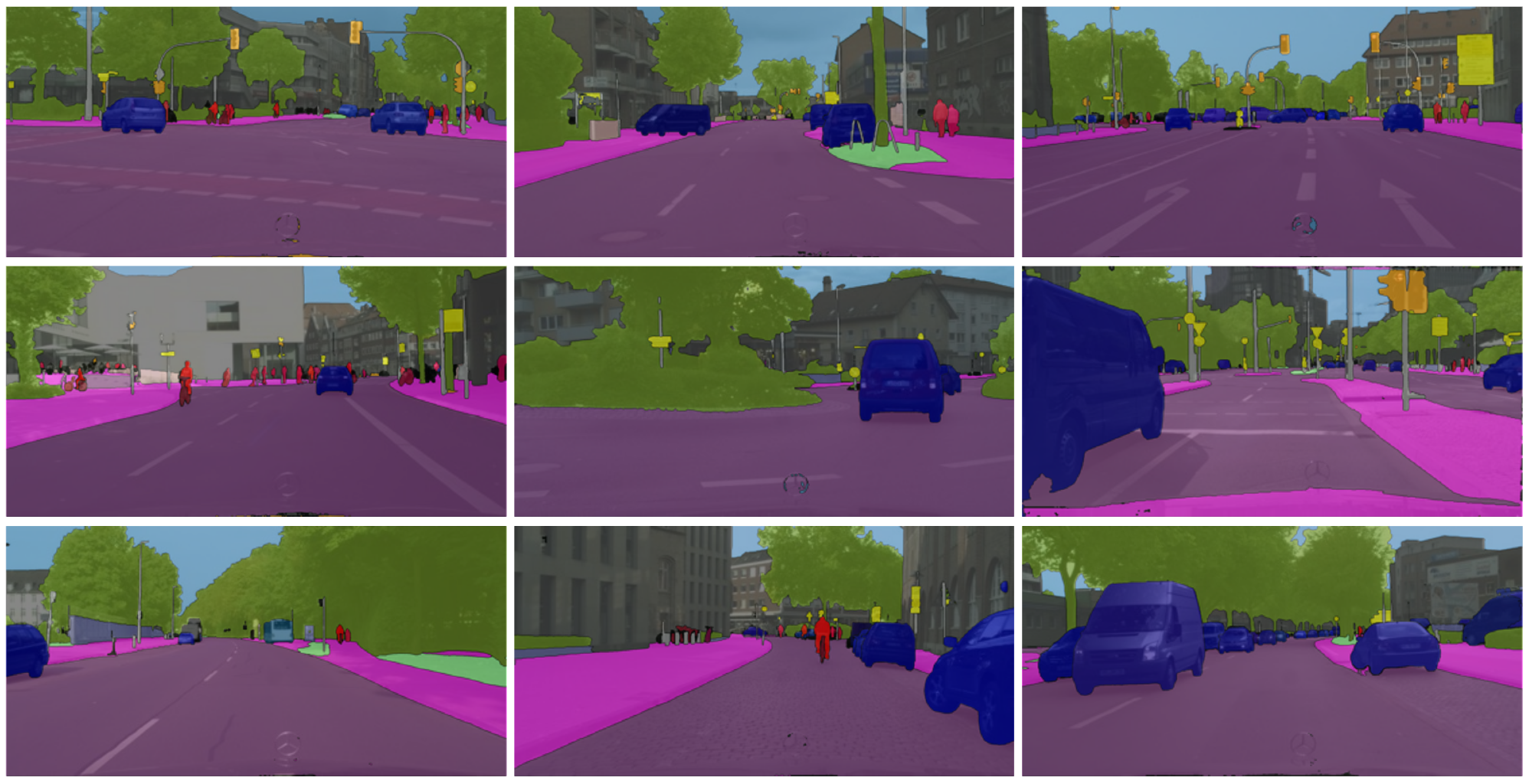} 
\caption{Visualization of panoptic results on the Cityscapes {\em val} set.
}
\label{fig:vis_city}
\end{figure*}

\vspace{0.5em}
\noindent
{\bf Cityscapes.} As for the street view, we visualize panoptic results on the Cityscapes {\em val} set, as illustrated in Fig.~\ref{fig:vis_city}. In addition to the well-depicted cars and pedestrians, the proposed approach achieves satisfactory performance on slender objects, like street lamps and traffic lights.

\begin{figure*}[t!]
\centering
\includegraphics[width=0.85\linewidth]{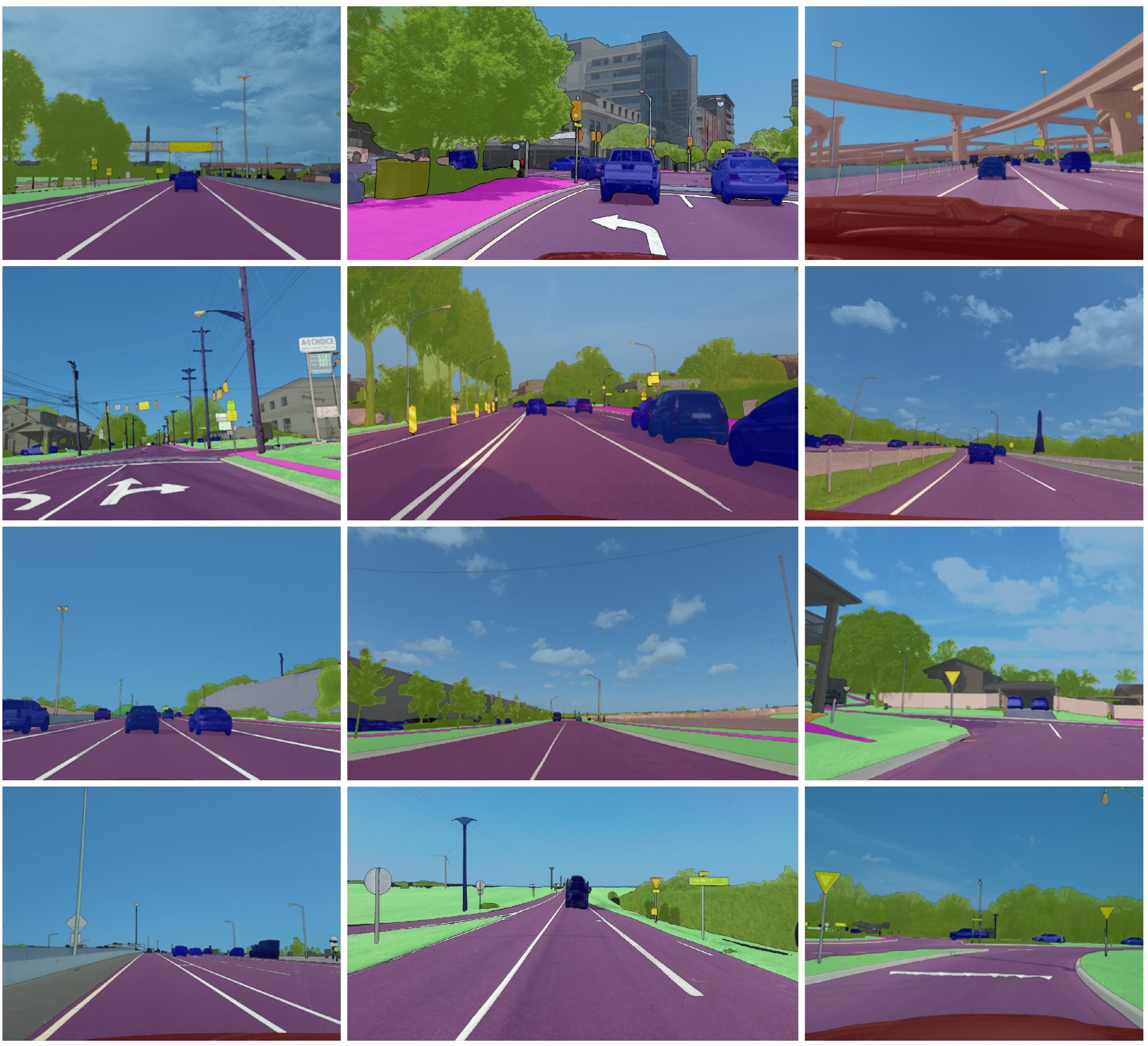} 
\caption{Visualization of panoptic results on the Mapillary Vistas {\em val} set.
}
\label{fig:vis_mapillary}
\end{figure*}

\vspace{0.5em}
\noindent
{\bf Mapillary Vistas.} In Fig.~\ref{fig:vis_mapillary}, we further present panoptic results on the Mapillary Vistas {\em val} set, which contains larger scales traffic-related scenes. It is clear in the figure that the proposed Panoptic FCN achieves surprising results, especially on vehicles and traffic signs. The coherence of qualitative results also reflects the fulfillment of {\em instance-awareness} and {\em semantic-consistency} in Panoptic FCN.

\end{document}